# SENTIMENT CLASSIFICATION of Customer's Reviews about Automobiles in Roman Urdu

Moin Khan, Kamran Malik

Punjab University College of Information Technology
University of the Punjab, Lahore, Pakistan
linktomoin@gmail.com, kamran.malik@pucit.edu.pk

*Abstract*—Text mining is a broad field having sentiment mining as its important constituent in which we try to deduce the behavior of people towards a specific item, merchandise, politics, sports, social media comments, review sites etc. Out of many issues in sentiment mining, analysis and classification, one major issue is that the reviews and comments can be in different languages like English, Arabic, Urdu etc. Handling each language according to its rules is a difficult task. A lot of research work has been done in English Language for sentiment analysis and classification but limited sentiment analysis work is being carried out on other regional languages like Arabic, Urdu and Hindi. In this paper, Waikato Environment for Knowledge Analysis (WEKA) is used as a platform to execute different classification models for text classification of Roman Urdu text. Reviews dataset has been scrapped from different automobiles' sites. These extracted Roman Urdu reviews, containing 1000 positive and 1000 negative reviews, are then saved in WEKA attribute-relation file format (arff) as labeled examples. Training is done on 80% of this data and rest of it is used for testing purpose which is done using different models and results are analyzed in each case. The results show that Multinomial Naïve Bayes outperformed Bagging, Deep Neural Network, Decision Tree, Random Forest, AdaBoost, k-NN and SVM Classifiers in terms of more accuracy, precision, recall and F-measure.

## I. INTRODUCTION

With the increase in computer usage and advancements in internet technology, people are now using their computers, laptops, smart-phones and tablets to access web and establishing their social networks, doing online businesses, e-commerce, e-surveys etc. They are now openly sharing their reviews, suggestions, comments and feedback about a particular thing, product, commodity, political affair and other viral news. Most of these are shared publicly and can be easily accessed from the web. Out of all those opinions, classifying the number of positive and negative opinions is a difficult task [1]. If you are planning to buy a particular product or choosing some institution, it's really difficult without any prior feedback regarding it. Similarly for the producers or service providers, it's a difficult ask for them to alter their SOPs without any review about their products or services from the customers. They can ask their customers to provide feedback via an e-survey, social media page or hand-written reviews. Their will so many opinions and reviews but their categorization as positive and negative is difficult. Some machine learning should be done to overcome this situation and to take betterment decisions later on.

The most important aspect in opinion mining is the sentiment judgment of the customer after extracting and analyzing their feedback. Growing availability of opinion-rich resources like online blogs, social media, review sites have raised new opportunities and challenges [2]. People can now easily access the publicly shared feedbacks and reviews which help in their decision making.

A lot of issues are involved in opinion mining. A major one is the handling of dual sense of the words, i.e. some words can depict a positive sense in a particular situation and a negative sense in the other. For example, the review: "the outer body of this car is **stiff**" shows positivity and hence the word **stiff** comes here in positive sense. Now consider another review "the steering wheel of this car is **stiff**" shows negativity and hence the word **stiff** interpreted here in negative sense [3]. Another issue is the understanding of sarcastic sentences e.g. "Why is it acceptable for you to be a bad driver but not acceptable for me to point out". One more issue is faced in the sentences which have both positive and negative meaning in them e.g. consider the sentences: "The only good thing about this car is its sporty look" and "Difficult roads often lead to beautiful destinations". Another commonly faced issue in opinion mining is the analysis of opinions, reviews and feedbacks shared on social media sites, blogs and review sites, which lack context and are often difficult to comprehend and categorize due to their briefness. Also they are mostly shared in the native language of the users; therefore to tackle each language according to its orientation is a challenging task (Rashid et al., 2013).

So far a lot of research work has been done on sentiment analysis in English language but limited work has been done for other languages being spoken around the globe. Urdu language, evolved in the medieval ages, is an Indo-Aryan language written in Arabic script and now had approx. 104 millions speakers around the globe [4]. Urdu can also be written in the Roman script but this representation does not have any specific standard for the correctness (correct spelling) of a word i.e. a same word can be written in different ways and with different spellings by different or even by the same person. Moreover, no one to one mapping between Urdu letters for vowel sounds and the Roman letters exist [5].

This research paper aims to mine the polarity of the public reviews specifically related to the automobiles and are written in Roman Urdu extracted from different automobiles review sites. The collected reviews dataset is used to train the machine using different classification models and then to assign the polarity of new reviews by using these trained classification models.

## II. RELATED WORK

Kaur et al. in 2014 [6] proposed a hybrid technique to classify Punjabi text. N-gram technique was used in combination with Naïve Bayesian in which the extracted features of N-gram model were supplied to Naïve Bayesian as training dataset; testing data was then supplied to test the accuracy of the model. The results showed that the accuracy of this model was better as compared to the existing methods.

Roman Urdu Opinion Mining System (RUOMiS) was proposed by Daud et al. in 2015, in which they suggested to find the polarity of a review using natural language processing technique. In this research, a dictionary was manually designed in order to make comparisons with the adjectives appearing in the reviews to find their polarity. Though the recall of relevant results was 100% but RUOMiS categorized about 21.1% falsely and precision was 27.1%.

Jebaseeli and Kirubakaran [7] proposed M-Learning system for the prediction of opinions as positive, negative and neutral using three classification algorithms namely Naïve Bayes, KNN and random forest for opinion mining. The efficiency of the stated algorithms was then analyzed using training dataset having 300 opinions equally split i.e. 100 opinions for each classification class (positive, negative and neutral). SVD technique was used in the preprocessing step to remove the commonly and rarely occurring words. 80% of the opinions were supplied as training dataset and rest was used for testing purpose. Highest accuracy was achieved by random forest classifier at around 60%.

Khushboo et al. presented opinion mining using the counting based approach in which positive and negative words were counted and then compared for the English language. Naïve Bayesian algorithm was used in this study. It is suggested that if the dictionary of positive and negative words is good, then really good results are returned. In order to increase the accuracy, change can be made in terms of parameters which are supplied to Naïve Bayesian algorithm.

Opinion mining in Chinese language using machine learning methods was done by Zhang et al. [8] using SVM, Naïve Bayes Multinomial and Decision Tree classifiers. Labeled corpus was trained using these classifiers and specific classification functions were learnt. The dataset comprises of Amazon China (Amazon CN) reviews. The best and satisfied results were achieved using SVM with string kernel.

A grammatical based model was developed by Syed et al. [9] which focused on grammatical structure of sentences as well as morphological structure of words; grammatical structures were extracted on basis of two of its substituent types, adjective phrases and nominal phrases. Further assignment was done by naming adjective phrases as Senti-Units and nominal phrases as their targets. A striking accuracy of 82.5% was achieved using shallow parsing and dependency parsing methods.

Pang et al. suggested the method to classify the documents on overall sentiment and not on the topic to determine the polarity of the review i.e. positive or negative. Movie reviews were supplied as dataset for training and testing purposes. It was found out that standard machine learning techniques performed pretty well and surpasses the human produced baselines. However, traditional topic-based categorization didn't produce good results using Naïve Bayes, maximum entropy classification and support vector machines.

Classification of opinions, using the sentiment analysis methodologies, posted on the web forum in Arabic and English languages was proposed by Abbasi et al. [10]. In this study, it was proposed that in order to handle the linguistic characteristics of Arabic language, certain feature extraction components should be used and integrated which returned very good accuracy of 93.62%. However, the limitation of this system was the domain specificness as only the sentiments related to hate and extremist groups' forums were classified by this system; because vocabulary of hate and extremist words is limited, so it isn't difficult to determine the polarity i.e. positive and negative words in the opinion.

## III. METHODOLOGY

The model proposed in this paper is divided into four main steps. First of all, the reviews written in Roman Urdu were scrapped from different automobiles sites [11], [12] as only automobiles reviews are targeted in this study. Training dataset was created and documented in text files using these scrapped reviews containing 800 positive and 800 negative reviews. Dataset is converted into the native format of WEKA, which is Attribute-Relation File Format (ARFF) and then that ARFF converted training dataset is loaded in the WEKA explorer mode to train the machine. Different models are developed by applying using different classifiers and then the results are analyzed and compared. The methodology comprises of the steps shown in figure 1.

### A. Material

The dataset used comprises of 2000 automobiles reviews in Roman Urdu with equal polarity of positive and negative reviews. 1600 example points are used for training the machine and the rest 400 are used for testing the accuracy of the models trained via different classifiers. This large training sentiment corpus was labeled prior to training the classification models.

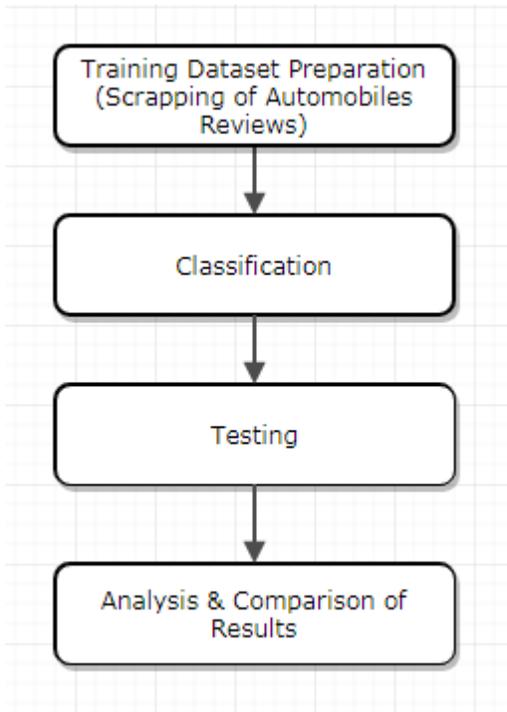

*Figure 1: Proposed model*

### B. Data Preprocessing

The purpose of this step is to ensure that only relevant features get selected from the dataset. In this step, before forwarding the data to the training of models and for classification, following steps were performed.

*1) Data Extraction*
The extraction task includes the scrapping of reviews from the automobiles sites. The users freely post their comments and reviews on these sites which are mostly multi-lingual, e.g. "Honda cars ka AC bohut acha hota hai", "imported cars k spare parts mehengy milty hn" etc. This is because English has a great influence on Urdu speaking community; and also due to the fact that most of the automobiles related terminologies is used as it is in other local languages as well including Urdu.

*2) Stop-words Removal*
Words which are non-semantic in nature are termed as stop-words and usually include prepositions, articles, conjunctions and pronouns. As they hold very little or no information about the sentiment of the review, so they are removed from the data [13], [14]. A list of Urdu stop-words was taken [15] and converted to Roman Urdu script.

*3) Lower-case words*
All the word tokens are converted to lower-case before they are added to the corpus in order to shift all the words to the same format, so that prediction can be made easy.

*4) Development of Corpus*
All the extracted reviews and comments are stored in a text file which includes 1000 positive and 1000 negative reviews. In this study, 800 positive and 800 negative reviews are used as training dataset and rest 400 reviews (200 positive and 200 negative) are used as testing dataset.

*5) Conversion of Data into ARFF*
Attribute-Relation File Format (ARFF) is an ASCII text input file format [16] of WEKA. These files have two important sections i.e. **Header** information and **Data** information. The dataset text file was converted to ARFF format using by using TextDirectoryLoader command in Simple CLI mode of WEKA. For example:
>java weka.core.converters.TextDirectoryLoader filename.txt > filename.arff
The elements of text files are saved as strings with relevant class labels.

### C. Classification

Before doing the classification part, all the string attributes are converted into set of attributes, depending on the word tokenizer, to represent word occurrence [17] using StringToWordVector filter. The set of attributes is determined by the training data.

Sentiment classification can be binary or multi-class sentiment classification. Binary classification has two polarities, e.g. good or bad, positive or neutral etc. In this paper, binary classification is done using machine learning algorithms. As training dataset is required for supervised machine learning algorithms, having feature vectors and their respective class labels, and a testing dataset [18]. A set of rules can be learnt by the classifiers from the training corpus prior to the testing process using the trained models. Multinomial Naïve Bayes, Bagging, Deep Neural Network, Decision Tree, Random Forest, AdaBoost, k-NN and SVM Classifiers are used to learn and classify the Roman Urdu dataset.

In WEKA, **Classify** tab is used to run different classifiers on the dataset. In this study, machine was trained using six classifiers and corresponding six models were built. These generated models are then used to predict about the testing data's polarity (positive or negative).

*1) Deep Neural Network*
Deep neural networks are Multi-Layer Perceptron Network. A number of neurons are connected to other neuron with the help of hidden layers. DNNs are also known as function approximator such as Fourier or Taylor. With enough layers of non-linear function approximator the DNNs can approximate any function. DNNs use non-linear function such as logistic, tan-hyperbolic etc. to compute the big fig function same with the principle of Fourier Series where sin and cos functions are used to determine the function. The coefficients in the DNN represent the same purpose such the coefficients of the Fourier series. As they have multiple layers in between input and output layers, complex non-linear relationships can be modeled using them.

*2) Decision Tree Classifier*

Decision Tree builds classification models in form of a tree like structure. The dataset is broken down into smaller chunks and gradually the corresponding decision tree is incrementally developed. The final outcome of this process depicts a tree with leaf nodes or decision nodes. It shows that if a specific sequence of events, outcomes or consequences is occurred, then which decision node has the maximum likelihood to occur and what class will be assigned to that sequence. The core logic behind decision trees is the ID3 algorithm. ID3 further uses Entropy and Information Gain to construct a DT.

*3) Bagging Classifier*

Bootstrap Aggregating Classifier is also known as the Bagging Classifier. This is a boosting and ensemble algorithm which combines the output of weak learners Decision Tree in most cases. It is known to have commonly used to avoid overfitting by reducing the variance. It creates m bootstrap samples which are used by classifiers and then the output is made by voting of the classifier.

*4) Random Forests*

Random Forests is also an ensemble algorithm which combines the output of decision trees. For regression it takes the mean of the output while for classification it goes for majority voting. It is remedy of decision tree's very common problem i.e. overfitting. It selects many bootstrap samples consisting of random features and thus is prone to lesser overfitting.

*5) Multinomial Naïve Bayes Classifier*

If discrete features like word counts for text classification are present, then the multinomial naïve bayes is very suitable. Integer feature count is normally required by the multinomial distribution. However, in practice, fractional counts such as tf-idf may also work.

*6) K-NN*

K Nearest Neighbor simply stores all available scenarios and then classifies new unseen scenarios using the similarity measure e.g. distance functions like Euclidean, Manhattan, Minkowski etc. It is a lazy learning technique learning technique as computation is deferred and approximation is done locally until classification. It can be used both for classification and regression purposes. In WEKA, it can be used by the name IBk algorithm.

*7) AdaBoost Algorithm*

Ada Boost is an adaptive boosting algorithm which is used to combine the output of weak learner algorithms. It works by tempering the misdiagnosed subsequent weak learners over previously misclassified records. This algorithm is sensitive to noise relatively is less prone to the overfitting. It works by calculating the weighted out of each learner in order to calculate the final output. Moreover, it only works on features that play predictive role and reduces the dimensionality of the data. It usually utilizes Decision tree as the weak learner. Decision tree with its own parameters can be set as a weak learner. The learning rate is compromised with the number of estimator trees.

*8) SVM*

A Support Vector Machine (SVM) is a discriminative classifier formally defined by a separating hyperplane. When given the labeled training data, SVM algorithm outputs an optimal boundary which classifies new cases. It uses kernel trick technique in order to transform the data and then finds an optimal boundary between the possible outputs based on the transformation which is done earlier.

*D. Testing of Models*

The testing dataset comprising of 200 positive and 200 negative reviews was supplied to the models which were trained on the training dataset. Same preprocessing steps were performed on the testing dataset and an ARFF file was gotten after executing these steps. It is then supplied to the WEKA by selecting the **Supplied test set** option in the Classify tab and then providing the test data ARFF file. After the file was loaded, each trained model was re-evaluated by right-clicking on each model and selecting the **Re-evaluate model on current test set option**. The results are noted down for all the classifiers.

*E. Results analysis and Comparison*

The classification results of all the classifiers used to classify the 400 new example points (testing dataset) and their accuracies are shown in table 1. The results of the testing process are analyzed and comparison is done among the classifiers in order to identify the classifier which best performed in the testing process and classified the testing reviews more accurately. The analysis is performed using the standard evaluation methodologies, i.e. precision, recall and F-measure (table 2). On the basis of these evaluations, classifiers are compared to declare the best one among all.

IV. CONCLUSION

In this study, we reviewed multiple text classification models to classify Roman Urdu reviews related to automobiles using Waikato Environment for Knowledge Analysis (WEKA). The dataset contained 1000 positive and 1000 negative. 80% of the reviews data was labeled and then for machine training, it was supplied to WEKA and different classification models were learnt. After that testing data was supplied and trained models were re-evaluated. The results showed that Multinomial Naïve Bayes performed best among all other classifiers in terms of more accuracy, precision, recall and F-measure (figure 2). For the computation part, Multinomial Naïve Bayes has better efficiency in learning and classification than the Decision Tree classifier [19] and consequently the classifiers which use decision tree at the backend. As far as our research is concerned, we have used Decision Tree, Bagging, Random Forests and AdaBoost classifiers which fall in the category of Decision Trees. The main reason of Multinomial Naïve Bayes being better in learning and classification than Decision Trees is that it shows a good probability estimate for correct class, which enables it to perform the correct classification [20].

| Classifier | Total Testing Reviews | Count of Correctly Classified Reviews | Count of Incorrectly Classified Reviews | Accuracy (%) |
|---|---|---|---|---|
| Deep Neural Network | 400 | 328 | 72 | 82 |
| Decision Tree | 400 | 303 | 97 | 75.75 |
| Bagging | 400 | 338 | 62 | 84.5 |
| Random Forests | 400 | 315 | 85 | 78.75 |
| Multi-nomial Naïve Bayes | 400 | 359 | 41 | 89.75 |
| k-NN | 400 | 288 | 112 | 72 |
| AdaBoost | 400 | 335 | 65 | 83.75 |
| SVM | 400 | 306 | 94 | 76.5 |

*Table 1: Accuracies of Classifiers*

| Classifier | Precision | Recall | F-Measure |
|---|---|---|---|
| Deep Neural Network | 0.88 | 0.92 | 0.9 |
| Decision Tree | 0.83 | 0.9 | 0.86 |
| Bagging | 0.89 | 0.94 | 0.91 |
| Random Forests | 0.85 | 0.95 | 0.9 |
| Multi-nomial Naïve Bayes | 0.93 | 0.96 | 0.95 |
| k-NN | 0.82 | 0.86 | 0.84 |
| AdaBoost | 0.91 | 0.92 | 0.92 |
| SVM | 0.87 | 0.87 | 0.87 |

*Table 2: Summarized Results of Evaluation Measures*

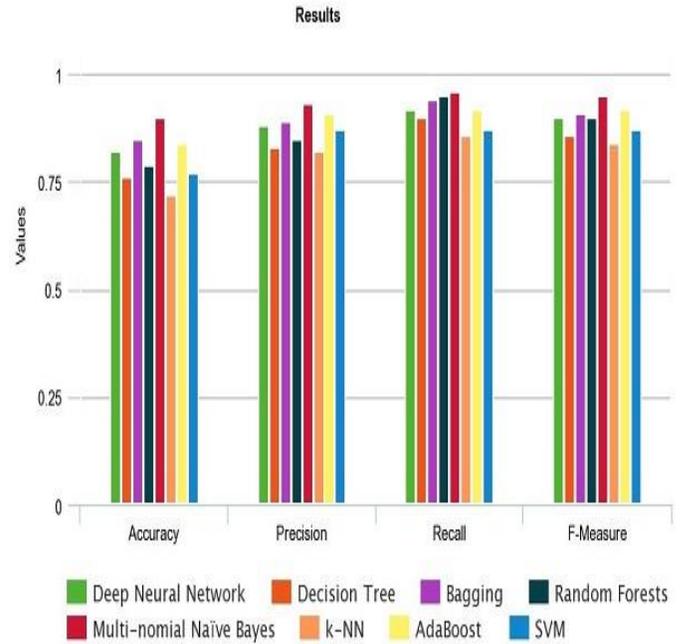

*Figure 2: Comparison of results of classifiers on testing dataset*

## V. FUTURE WORK

In this study, we have targeted the Roman Urdu reviews related to automobiles; this approach can be extended and can be used for other fields as well like hotels reviews, taxi service reviews that are provided in Roman Urdu.

We can train our machine to guess the exact area of interest from the given Roman Urdu review, e.g. if a person is talking about the engine of the automobile, our system should point that out.

Many reviews are shared with neutral polarity. Currently we have done this research to handle binary sentiment classification of the reviews i.e. positive and negative. This can be extended to handle multi-class sentiment classification; in this way, we can handle reviews having neutral polarity as well.